\documentclass{article}
\usepackage{spconf,amsmath,graphicx}
\usepackage{graphicx,psfrag,epsf}
\usepackage{enumerate}
\usepackage{bbold}
\usepackage{multirow,array}
\usepackage{changepage}
\usepackage{verbatim}

\usepackage{hyperref}
\usepackage{epsfig,bm}
\usepackage{epstopdf}
\usepackage{color}

\usepackage{amsfonts}
\usepackage{bm}
\usepackage{amssymb}
\usepackage{float}
\usepackage[utf8]{inputenc}
\usepackage[english]{babel}
\usepackage[ruled,vlined,linesnumbered]{algorithm2e}

\title{B-SMALL: A Bayesian Neural Network Approach to Sparse Model-Agnostic Meta-learning}

\name{Anish Madan, Ranjitha Prasad}
\address{Indraprastha Institute of Information Technology Delhi, New Delhi}

\begin{document}

\maketitle
\vspace{-10mm}

\begin{abstract}
There is a growing interest in the learning-to-learn paradigm, also known as meta-learning, where models infer on new tasks using a few training examples. Recently, meta-learning based methods have been widely used in few-shot classification, regression, reinforcement learning, and domain adaptation. The model-agnostic meta-learning (MAML) algorithm is a well-known algorithm that obtains model parameter initialization at meta-training phase. In the meta-test phase, this initialization is rapidly adapted to new tasks by using gradient descent. However, meta-learning models are prone to overfitting since there are insufficient training tasks resulting in over-parameterized models with poor generalization performance for unseen tasks. In this paper, we propose a Bayesian neural network based MAML algorithm, which we refer to as the \texttt{B-SMALL} algorithm. The proposed framework incorporates a sparse variational loss term alongside the loss function of MAML, which uses a sparsifying approximated KL divergence as a regularizer. We demonstrate the performance of \texttt{B-MAML} using classification and regression tasks, and highlight that training a sparsifying BNN using MAML indeed improves the parameter footprint of the model while performing at par or even outperforming the MAML approach. We also illustrate applicability of our approach in distributed sensor networks, where sparsity and meta-learning can be beneficial.
\end{abstract}
\vspace{-5mm}
\section{Introduction}
\label{sec:intro}

The ability to adapt and learn new models with small amounts of data is a critical aspect of several systems such as IOTs, secure communication networks, biomedical signal processing, image processing etc. Traditional signal processing has addressed such problems using Bayesian and sparse signal processing techniques under a model-driven approach, incorporating several statistical assumptions on the distribution of input data. However, the modern era of artificial intelligence, brings in the promise of model-free processing using various machine learning algorithms, with no assumptions required on the statistical properties of the signals involved. 

Among several machine learning approaches proposed to deal with low-data regimes \cite{olson2018modern,lake2015human}, meta-learning is a simple yet efficient technique which aims at obtaining rapid adaptation across various \emph{tasks}\footnote{Often, \emph{task} refers to a subset of observations sampled from the original dataset in such a way that only a subset of the final prediction problem can be solved in the task.}, given small amount of data for updating the parameters pertaining to each task. In particular, model-agnostic meta-learning (MAML) is an algorithm that trains a model’s parameters such that a small number of gradient updates will lead to fast learning on a new task. Specifically, MAML obtains a meta-initialization at meta-training phase using task-specific training, and this initialization is rapidly adapted to a new task by using gradient descent in the meta-test phase. MAML is a baseline for any state-of-the-art few-shot learning method since it has been used for supervised classification, regression and reinforcement learning in the presence of task variability. Furthermore, MAML substantially outperforms techniques that use pre-training as initialization. In order to further improve on the adaptation and accuracy performance of MAML, several authors have proposed modifications such as introducing novel regularizers by analysing the optimization landscape \cite{guiroy2019towards}, feature reuse perspective based ANIL framework \cite{raghu2019rapid}, a meta-regularizer using information theoretic approaches for mitigating the memorization problem in MAML \cite{yin2019meta}, etc.

In signal processing based applications, such as distributed signal processing, there is a need for a technique that rapidly adapts in a distributed manner using low amount of heterogeneous data at each sensor node. Furthermore, it is essential that these machine learning models be computationally simple and memory-efficient in terms of the number of parameters they require \cite{dutta2020discrepancy}. The inherent structure of the MAML algorithm lends itself in such scenarios since the task level learning in the inner iteration can be associated to per-node learning, while outer iteration parameter update agglomerates the updates from neighboring nodes, effectively enabling inference capability at each node. However, a challenge in the existing meta-learning approaches is their tendency to overfit, thereby defeating the true purpose of designing such networks \cite{yoon2018bayesian}. It is well-known that incorporating sparsity constraints during model training guarantees statistical efficiency and robustness to overfitting, hence improving generalization performance on previously unseen tasks \cite{tipping2001sparse}. In the context of compact meta-learning, network pruning \cite{tian2020meta} and regularization \cite{tseng2020regularizing,gai2019sparse} have led to sparse meta-learned models without compromising generalization performance. Several methods have been proposed in order to combine deep networks and probabilistic methods for few-shot learning. In particular, in  \cite{edwards2016towards}, the authors employ hierarchical Bayesian models for few shot learning. In \cite{finn2018probabilistic}, the authors employ  a graphical model via a hierarchical Bayesian model that includes prior distribution over the weights and hyperparameters of the meta-learning model. 

A popular approach for incorporating uncertainty in deep networks is using the \emph{Bayesian neural networks} (BNN) \cite{blundell2015weight, kingma2015variational}. Although exact inference in BNNs is not possible \cite{hron2018variational}, approximations based on backpropagation and sampling have been effective in incorporating uncertainty into the weights \cite{kingma2015variational}. Furthermore, these networks can be made sparse, and eventually compressed to obtain light neural networks \cite{dmitry2017variational}. However, so far, conventional BNNs directly learn only the posterior weight distribution for a single task and have not been employed in the meta-learning framework. 

\noindent \textbf{Contributions}: We build a meta-learning based method to tackle low-data based ambiguity that occurs while learning from small amounts of data using simple albeit highly-expressive function approximators such as neural networks. To enable this, our natural choice is the optimization based MAML framework. In order to abate overfitting we propose to design a sparse MAML algorithm for BNNs. We propose \texttt{B-SMALL}, where, in each of the parameter update steps of the MAML algorithm, the parameters of the BNN are updated using the \emph{sparse} variational loss function proposed in the context of the sparse variational dropout (SVD) algorithm \cite{dmitry2017variational}.
 
We demonstrate the effectiveness of this technique in achieving sparse models with improved accuracy on well-known datasets in the context of classification as well as regression\footnote{Code for the experiments can be found on github at \href{https://github.com/anishmadan23/B-SMALL}{https://github.com/anishmadan23/B-SMALL}}. Finally, we present a use-case for the proposed \texttt{B-SMALL} algorithm as distributed algorithms for sensor networks. 
\vspace{-5mm}
\section{Preliminaries}
In this section, we describe MAML, an optimization based meta-learning paradigm, followed by description of the Bayesian neural network and the SVD paradigm. 

\subsection{Model-Agnostic Meta-Learning (MAML)}

MAML considers a set of tasks distributed as $p(\mathcal{T})$, for few-shot meta-learning. In a given meta-training epoch, a model represented by a parameterized function $f_\theta$ with parameters $\theta$ is adapted to a new task $\mathcal{T}_i$ drawn from $p(\mathcal{T})$, using $K$ samples drawn from the data distribution($K$-shot). The resulting  update (single) of the model’s parameters given by
\begin{equation}
    \bm{\theta'}_i = \bm{\theta} - \gamma \nabla_{\bm{\theta}} \mathcal{L}^{\mathcal{T}_i}(f_{\bm{\theta}}).
    \label{eq:MAMLInner}
\end{equation}
Typically, the parameter updates are computed using a few gradient descent steps evaluated for each task $\mathcal{T}_i$. The outer iteration consists of meta-optimization across all the tasks, the model parameters are updated using as given by 
\begin{equation}
    \bm{\theta} \leftarrow \bm{\theta} - \beta \nabla_{\bm{\theta}} \sum_{\mathcal{T}_i \sim p(\mathcal{T})} \mathcal{L}^{\mathcal{T}_i}(f_{\bm{\theta'_i}}),
    \label{eq:MAMLInnerOuter}
\end{equation}
where $\beta$ is the meta step-size. Hence, the test error on sampled tasks $\mathcal{T}_i$ is the training error of the meta-learning process~\cite{finn2017model}. 

\subsection{Bayesian Neural Networks and Sparsity}

Among several manifestations of employing Bayesian methods in deep neural networks, Bayesian inference based on variational dropout (VD) for inferring the posterior distribution of network weights is quite popular \cite{kingma2015variational} . In \cite{dmitry2017variational}, the authors proposed the sparse variational dropout (SVD) technique where they provided a novel approximation of the KL-divergence term in the VD objective \cite{kingma2015variational}, and showed that this leads to sparse weight matrices in fully-connected and convolutional layers. The resulting BNNs are robust to over-fitting, learn from small datasets and offer uncertainty estimates through the parameters of per-weight probability distributions. 

Consider a BNN with weights $\mathbf{w}$, and a prior distribution over the weights, $p(\mathbf{w})$. Training a BNN involves optimizing a variational lower bound given by
\begin{align}
\mathcal{L}(\phi) =  \mathcal{L}_\mathcal{D}(\phi)- D_{KL}(q_\phi(\mathbf{w}) \Vert p(\mathbf{w})),
\label{eq:elbo}
\end{align}
where $\mathcal{L}_\mathcal{D}(\phi) = \mathbb{E}_{q_\phi(\mathbf{w})}[\log(p(y_n|x_n,\mathbf{w}))]$, $q_\phi(\mathbf{w})$ is an approximation of the true posterior of the weights of the network parameterized by $\phi$, and $D_{KL}(q_\phi(\mathbf{w}) \Vert p(\mathbf{w}))$ is the KL-divergence between the true posterior and its approximation. We employ the approximation of the above variational lower bound, termed as sparse variational dropout, as derived in \cite{dmitry2017variational}. Here, a multiplicative Gaussian noise $\zeta_{i,j} \sim \mathcal{N}(1, \alpha)$ is applied on a weight $w_{i,j}$, which is equivalent to sampling $w_{i,j}$ from $\mathcal{N}(w_{i,j}| \theta_{i,j} , \alpha_{i,j} \theta_{i,j}^2)$. Training the BNN involves learning $\alpha_{i,j}$, and $\theta_{i,j}$, i.e., the variational parameters are given by $\phi_{i,j} = [\alpha_{i,j},\theta_{i,j}]$.

\vspace{-3mm}
\section{BNN based Sparse MAML (\texttt{B-SMALL})}
Consider a task distribution $p(\mathcal{T})$ over the set of tasks $\mathcal{T}$ that encompasses the data points in $\mathcal{D} = \{x_j, y_j \}$, for $j = 1, \hdots, N$. These tasks $\mathcal{T}_i \in \mathcal{T}$ are used for meta-training a  model $p(y|x,\mathbf{w})$ which predicts $\mathbf{y}$ given inputs $\mathbf{x}$ and parameters $\mathbf{w}$. We adopt the SVD framework to introduce sparsity and reduce overfitting in our meta-learning pipeline, i.e., we maximize the variational lower bound and accordingly modify the loss function of MAML in the inner and outer loop as follows:
\begin{align}
     \mathcal{L}^{T_i}(\phi) &=  \mathcal{L}_{D}^{T_i}(\phi) -  D_{KL}[q_\phi(\mathbf{w})|| p(\mathbf{w})].
\label{eq:main_eq}
\end{align}
Here, similar to the previous section, $\phi$ denotes the variational parameters given by $\phi=[\theta,\alpha]$. Further,  $\alpha$ is interpreted as the dropout rate. Note that SVD enables us to have individual dropout rates for each neuron that are learnable. Furthermore, the regularization term is such that $\alpha_{i,j} \rightarrow +\infty$ for several neurons. A high dropout value implies we can effectively ignore the corresponding weight or neuron and remove it from the model, leading to lighter neural network models. Also, $\mathcal{L}^{D}_{T_i}(\phi)$ takes the form similar to the cross entropy loss for discrete classification problem, and squared loss in the case of regression problem \cite{dmitry2017variational}.

\begin{algorithm}[h]
\SetAlgoLined
  \textbf{Parameters }: $\phi = [\theta,\alpha] $ \\ 
  Initialize $\phi$  \\  

  \textbf{Hyperparams}: $\gamma, \beta$ (Step-size) \\

  \While{not done}{
    Sample batch of tasks $\mathcal{T}_i \sim \mathcal{T}$ \\
     \For{all $\mathcal{T}_i$}{
         Sample $K$ points $\mathcal{D}_i=\{x_{(k)},y_{(k)}\}$ from $\mathcal{T}_i $ \\
        Evaluate $\nabla_\phi \mathcal{L}_{\mathcal{T}_i}(\phi)$ using $\mathcal{D}_i$ w.r.t. \eqref{eq:main_eq} \\
        Compute $\phi_{i}'$ as in \eqref{eq:MAMLInner} using $\mathcal{D}_i' $\\
     }
     Update $\phi \leftarrow \phi - \beta {\nabla}_{\phi} \sum_{ {\mathcal{T}}_i \sim p(\mathcal{T}) } \mathcal{L}_{\mathcal{T}_i}(\phi_{i}') $
  }
  \caption{B-SMALL Algorithm}
  \label{training_algo}
 \end{algorithm}

\vspace{-5mm}
\section{Experiments and Results}

In this section, we illustrate the performance of the proposed \texttt{B-SMALL} approach in the context of both, classification and regression. We evaluate the classification performance on few-shot image recognition benchmarks such as the Mini-Imagenet\cite{ravi2016optimization} and CIFAR-FS\cite{oreshkin2018tadam} datasets \cite{finn2017model}. The setup is a $N$-way, $K$-shot experiment, where we randomly select $N$ classes and choose $K$ images/samples for each class at every training step. All the models in different experiments are trained for $60000$ steps. We measure sparsity as the ratio of total number of weights above a certain threshold $\eta$, and total number of  weights. We set $\eta = 3$ for all our experiments, and consider those neurons as dropped out when $\log \alpha_{i,j} > \eta$, where $\alpha_{i,j}$ is the variational parameters in \eqref{eq:main_eq}.
\vspace{-3mm}
\subsection{$K$-Shot Regression}
\begin{figure}[htb!]
    \centering
    \includegraphics[width=0.9\linewidth]{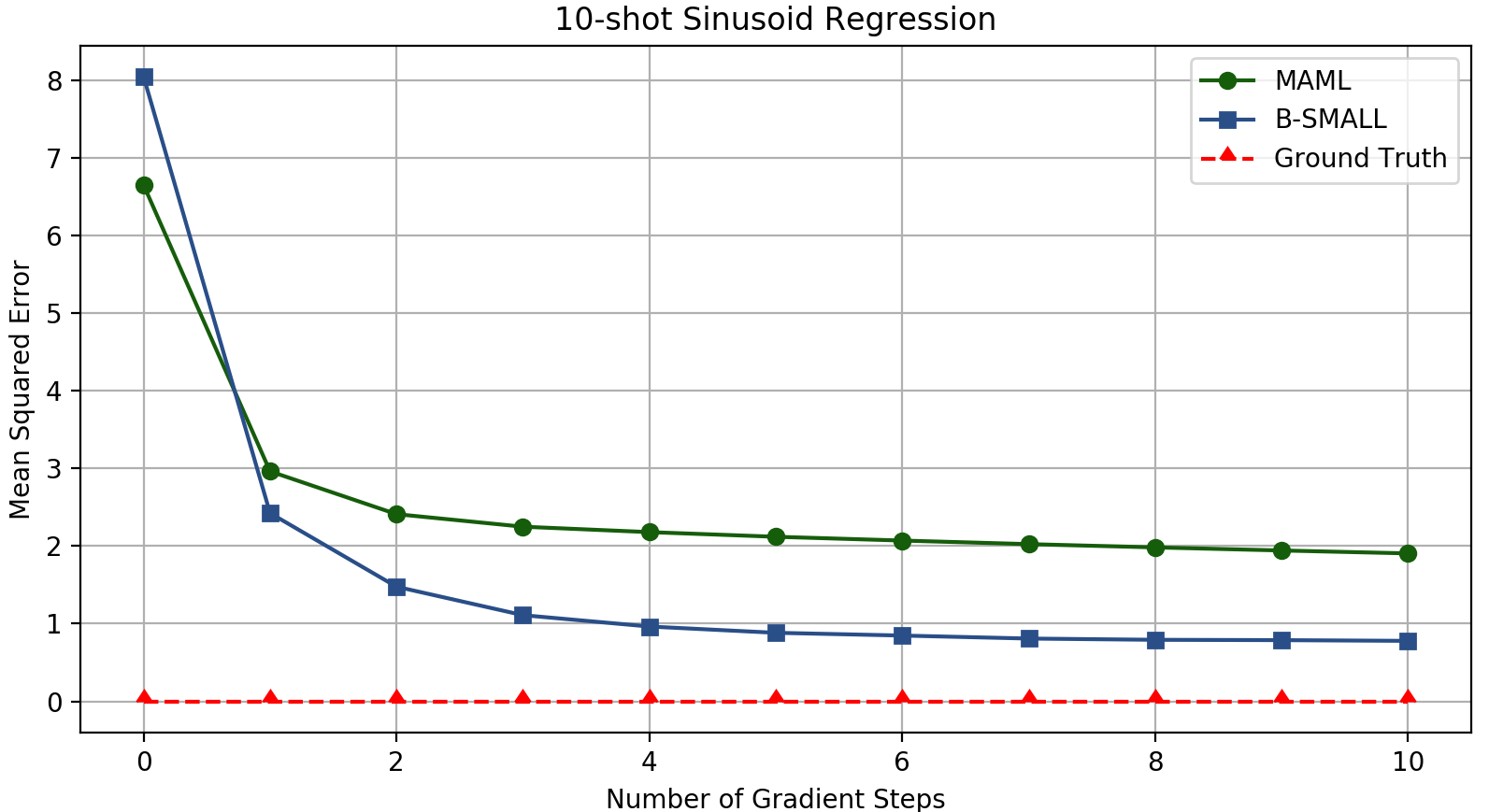}
    \caption{Plot of MSE vs Number of Gradient Steps taken at meta-test time for $K=10$ Sinusoid Regression.}
    \label{fig:mse_sine}
\end{figure}

We illustrate the performance of the proposed \texttt{B-SMALL} framework on $K$-shot regression, where the underlying ground-truth function that relates the input to the output is $sin(\cdot)$. We choose the amplitude range as $[0.1,5]$ and phase as $[0,\pi]$ and construct the meta-train and meta-test sets by sampling data points uniformly from $[-5.0,5.0]$. We choose a neural network with $2$ hidden layers of $40$ neurons each, followed by ReLU activation for this experiment. Further, we train the meta-learning model using a single gradient step, and a fixed step size $\gamma = 0.01$. We train the models only for $K=10$ and fine tune it for $K=\{5,20\}$. We evaluate mean square error (MSE) for $600$ random test points, all of which are adapted using the same $K$ points. The results in Table~\ref{tab:sine_reg} are averaged over $3$ different random seeds. We note that like MAML, \texttt{B-SMALL} also continues to improve after a single gradient step (i.e., the number of gradient steps it was trained on as depicted in Fig.~\ref{fig:mse_sine}). This implies that \texttt{B-SMALL} is able to find an initialization for the model such that it lies in region where further improvement is possible, while providing better MSE scores when compared to MAML, as depicted in Table~\ref{tab:sine_reg}. Furthermore, \texttt{B-SMALL} outperforms MAML in all 3 cases alongside providing sparse weight matrices. Even on such a small model, we manage to get $18\%-27\%$ sparsity.

\vspace{-3mm}
\subsection{Few-Shot Classification}

\begin{figure}
    \centering
    \includegraphics[width=0.9\linewidth]{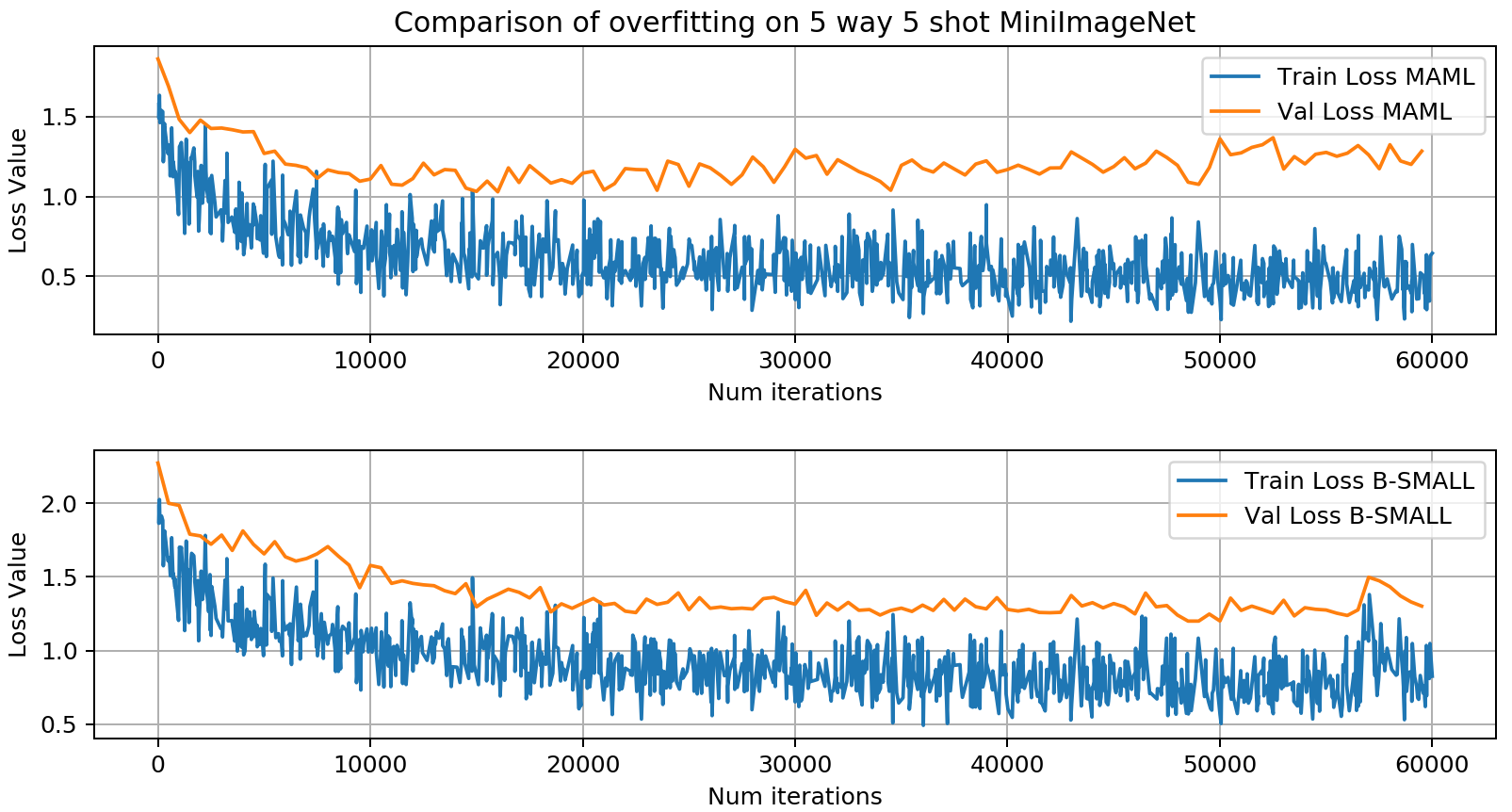}
    \caption{Overfitting in the case of MAML and B-SMALL: Note that difference between train and validation loss for MAML is much higher than that for B-SMALL, thereby showing the effect of regularization and enabling better learning.}
    \label{fig:overfit}
\end{figure}

To illustrate the few-shot classification performance of \texttt{B-SMALL}, we use the Mini-Imagenet dataset which consists of $100$ classes from the Imagenet dataset \cite{krizhevsky2012imagenet}, with each class containing $600$ images, resized to $84\times 84$ for training. The dataset is divided into $64$ classes for training, $16$ for validation and $20$ for testing. We also use the CIFAR-FS dataset proposed in \cite{oreshkin2018tadam}, which consists of $100$ classes and follows a similar split as Mini-Imagenet. We use a neural network architecture with $4$ blocks, where each block contains $ 3\times 3 $ Convolution, batch normalization, a ReLU layer \cite{finn2017model}. We also use a Maxpooling layer with kernel size $2 \times 2$, which is useful to reduce the spatial dimensionality of the intermediate features. We use $32$ filters for each convolutional layer. The models were trained using $5$ gradient steps, with step size $\gamma = 0.01$, and evaluated them using $10$ steps. We use a batch size of $4$ and $2$ tasks for $5$ and $1$-shot training, respectively. 
We observe that \texttt{B-SMALL} performs on par with or outperforms  MAML as depicted in Table.~\ref{mini} and Table.~\ref{cifarfs}. The aspect to be highlighted is that \texttt{B-SMALL} leads to sparse models which enables less overfitting during meta-train as depicted in Fig.~\ref{fig:overfit}. An interesting observation is the amount of sparsity in each case - when input information is large (more examples while training, i.e., higher $K$ in $K$-shot), the models are less sparse since the network encodes the additional information into its weights, in order to drive its decisions. 

\begin{table}[h]
    \centering
    \begin{tabular}{|c|c|c|}
    \hline
    {Model/Experiment} & \multicolumn{2}{c|}{5 way accuracy}\\
    
    & 1 shot & 5 shot \\
    \hline

    MAML\cite{finn2017model} & $48.70 \pm 1.84\% $ & $63.11 \pm 0.92\% $ \\
    CAVIA\cite{zintgraf2019fast} & $47.24 \pm 0.65\%$ & $61.87 \pm 0.93\%$ \\
    \hline
    MAML(Ours) & $46.30 \pm 0.29\%$ 
    & $66.3 \pm 0.21\%$ \\
    \hline
    B-SMALL & $\mathbf{49.12 \pm 0.30\%} $ & $\mathbf{66.97 \pm 0.3\% }$\\
    
    Sparsity & $\mathbf{76\%}$ & $\mathbf{44\%}$ \\

    \hline
\end{tabular}
\caption{Few-shot classification results on the Mini-Imagenet Dataset. The $\pm$ shows 95\% confidence interval over tasks. We compare it with our baseline MAML\cite{finn2017model} and CAVIA\cite{zintgraf2019fast} as reported in their papers. We include CAVIA as it improves on MAML by reducing overfitting. Additionally we also implement MAML (i.e MAML(Ours)) to ensure results are comparable to those reported.}
\label{mini}
\end{table}

\begin{table}[h]
    \centering
    \begin{tabular}{|c|c|c|}
    \hline
    {Model/Experiment} & \multicolumn{2}{c|}{5 way accuracy}\\
    
     & 1-shot & 5-shot\\
    \hline
    MAML\cite{bertinetto2018meta} & $58.9 \pm 1.9\%$ & $71.5 \pm 1.0\%$ \\
    \hline
    MAML(Ours) & $59.3 \pm 0.25\%$  & $70.85 \pm 0.19\%$\\
    \hline
    B-SMALL & $\mathbf{59.8 \pm 0.29\%}$ & $67.53 \pm 0.25\% $\\
    Sparsity & $\mathbf{37\%} $ & $\mathbf{34\%}$ \\
    \hline \hline
    {Model/Experiment} & \multicolumn{2}{c|}{2 way accuracy} \\
    & 1-shot & 5-shot \\
    \hline
    MAML\cite{bertinetto2018meta} &  $ 82.8 \pm2.7\%$ & $88.3 \pm 1.1\%$\\
    \hline
    MAML(Ours) & $ 80.20 \pm 0.26\%$ & $ 88.43 \pm 0.4\% $  \\    
    \hline
    B-SMALL & $ \mathbf{85.06 \pm 0.28\%}$ & $\mathbf{88.96 \pm 0.24\%} $  \\ 
    Sparsity & $ \mathbf{46\%} $ & $ \mathbf{45\% }$ \\

    \hline
\end{tabular}
\caption{Few-shot classification on CIFAR-FS Dataset.}
\label{cifarfs}
\end{table}

\begin{table}[h]
    \centering
    \begin{tabular}{|c|c|c|c|}
    \hline
    Expt &  \multicolumn{3}{c|}{MSE @ Num Grad Steps} \\
    & 1 & 5 & 10 \\
    \hline
    MAML (k=5) & 0.8347 & 0.5415 & 0.5668 \\
    B-SMALL (k=5) & \textbf{0.7697} & \textbf{0.4596} &  \textbf{0.4392} \\
    \hline
    MAML (k=10) & 1.493 & 0.8088&  0.7119 \\
    B-SMALL (k=10) & \textbf{1.2007} & \textbf{0.3816} &  \textbf{0.3386}\\
    \hline
    MAML (k=20) & 0.5238 & 0.0848&  \textbf{0.04555} \\
    B-SMALL (k=20) & \textbf{0.3445} & \textbf{0.0628} &  0.0518\\
    \hline
\end{tabular}
\caption{MSE for $K$-shot sinusoid regression: MAML Vs. \texttt{B-MAML} at gradient steps $\{1,5,10\}$ after training with a single gradient step.}
\label{tab:sine_reg}
\end{table}

\vspace{-3mm}
\subsection{Use-case: Sparse MAML in Sensor Networks}

Consider a sensor network whose communication links are governed by a graph given by $\mathcal{G} = (\mathcal{V},\mathcal{E})$, where $\mathcal{V}$ represents the set of vertices with $|\mathcal{V}| = V$ and $\mathcal{E}$ represents the set of edges. The degree of the $i$-th vertex is given by $\mathcal{D}(i)$, and each vertex is equipped with a neural network. We also assume that the sensors are connected to fusion center, which can communicate with all the sensor nodes. Without loss of generality, we assume that at the $i$-th vertex, the neural network learns the model parameters pertaining to a set of tasks, $\mathcal{T}_i \in \mathcal{T}$. Translating the MAML algorithm for the sensor network, say the inner iterations of the MAML algorithm are executed at each node, i.e., at the $i$-th node, the parameter update is given by \eqref{eq:MAMLInner}. The inner iteration update is communicated to the fusion center, which obtains such updates from other vertices as well. The fusion center executes the outer iteration using \eqref{eq:MAMLInnerOuter}, and the final updated weights can be communicated to the sensors for the purposes of inference.
It is challenging to extend \texttt{B-MAML} to a distributed sensor network (in the absence of the fusion center). For instance, if $\mathcal{G}$ is a complete graph, i.e., $\mathcal{D}(i) = V-1$ for all $i$, then it is possible to implement exact MAML with a slight modification to the original algorithm. Furthermore, individual sensors have limited computational capability, and  bandwidth of the communication links are limited. Hence, it is pertinent to develop distributed algorithms that are memory-efficient with minimal message exchange between nodes. We address these aspects of \texttt{B-SMALL} in future work.
\vspace{-5mm}
\section{Conclusion and Future Work}

In this paper, we proposed the \texttt{B-SMALL} framework, a sparse BNN-based MAML approach for rapid adaptation to various tasks using small amounts of data. The parameters of the BNN are learnt using a sparse variational loss function. We demonstrated that the proposed framework outperformed MAML in most of the scenarios, while resulting in sparse neural network models. The results obtained builds on the theory that often, in deep learning, we have more parameters as compared to training instances, and such models are prone to overfitting \cite{hinton2012improving}. This gap is amplified in meta-learning since it operates in the low-data-regime and therefore it is important to use regularization technique as in \texttt{B-SMALL}. This helps to reduce the parameter footprint thereby reducing overfitting, and boosts generalization performance. As a future work, we plan to design and analyse \texttt{B-MAML} type algorithms for distributed processing.  

\vspace{-5mm}
\bibliographystyle{IEEEbib}
\bibliography{MetaLearnBayesianDistributed.bib}

\end{document}